\documentclass[letterpaper]{article}
\usepackage{uai2019}
\usepackage[margin=1in]{geometry}
\usepackage{times}

\usepackage{booktabs}
\usepackage{amsthm, amssymb, amsmath}
\usepackage{mathtools}
\usepackage{graphicx}
\usepackage{algorithm}
\usepackage{algorithmicx}
\usepackage[noend]{algpseudocode}
\usepackage{todonotes}

\usepackage{stmaryrd, color, tikz, dsfont}
\usepackage{subcaption}
\usepackage{booktabs}

\usepackage[hidelinks]{hyperref}

\usepackage{natbib}
\newcommand{\dbn}{{\sc dbn}}
\newcommand{\dbns}{{\sc dbn}s}
\newcommand{\mdp}{{\sc mdp}}
\newcommand{\fmdp}{{\sc fmdp}}

\newcommand{\pomdp}{{\sc pomdp}}
\newcommand{\Pa}{\mathit{Pa}}
\newcommand{\iti}{{\sc iti}}
\newcommand{\svi}{{\sc svi}}

\newtheorem{theorem}{Theorem}
\newtheorem{lemma}{Lemma}
\newtheorem{definition}{Definition}

\title{Learning Factored Markov Decision Processes with Unawareness}		

\author{ 
	{\bf Craig Innes} \\
	School of Informatics \\
	University of Edinburgh\\
	Edinburgh, United Kingdom EH8 9AB \\
	craig.innes@ed.ac.uk
	\And
	{\bf Alex Lascarides}  \\
	School of Informatics \\
	University of Edinburgh \\
	Edinburgh, United Kingdom EH8 9AB \\
	alex@inf.ed.ac.uk
}

\begin{document}
	
	\maketitle
	
	\begin{abstract}
		Methods for learning and planning in sequential decision problems often assume the learner is aware of all possible states and actions in advance. This assumption is sometimes untenable. In this paper, we give a method to learn factored markov decision problems from both domain exploration and expert assistance, which guarantees convergence to near-optimal behaviour, even when the agent begins unaware of factors critical to success. Our experiments show our agent learns optimal behaviour on small and large problems, and that conserving information on discovering new possibilities results in faster convergence.
	\end{abstract}
	
	\section{Introduction}
	\label{sec:introduction}
	
	Factored markov decision processes (\fmdp{}s) are a fundamental tool for modelling complex sequential problems. When the transition and reward functions of an \fmdp{} are known in advance, there are tractable methods to learn its optimal policy via dynamic programming \citep{guestrin_efficient_2003}. When these components are unknown, methods exist to jointly learn a structured model of the transition and reward functions \citep{degris_learning_2006,araya-lopez_active_2011}. Yet all such methods (with the exception of \cite{rong_learning_2016}) assume that the way the domain of the problem is \emph{conceptualized}---the possible actions available to the agent and the belief variables that describe the state space---are completely known in advance of learning. In many scenarios, this assumption does not hold.
	
	For example in medicine, suppose an agent prescribes a particular drug, but later a senior pharmacologist objects to the prescription based on a reason unforeseen by the agent---the patient carries a newly discovered genetic trait, and the drug produces harmful side effects in its carriers. Further, this discovery may occur after the agent has already learned a lot about how other (foreseen) factors impact the drug's effectiveness. As \cite{coenen_asking_2017} point out, such scenarios are common in human discussion---the answer to a person's inquiry may not only provide information about which of the questioner's existing hypotheses are likely, but may also reveal entirely new hypotheses not yet considered. This example also shows that while it may be infeasible for an agent to gather all relevant factors of a problem \emph{before} learning, it may be easy for an expert to offer contextually relevant corrective advice \emph{during} learning. Another example is in robotic skill learning. Methods such as \cite{cakmak_designing_2012} enable an expert to teach a robot \emph{how} to perform a new action, but don't teach \emph{when} it's optimal to use it. In lifelong-learning scenarios, we want to integrate new skills into existing decision tasks without forcing the robot to restart learning each time.
	
	Current models of learning and decision making don't address these issues; they assume the task is to use data to \emph{refine} a distribution over a \emph{fixed} hypothesis space. Under this framework, any change to the set of possible hypotheses constitutes an unrelated problem. The above examples, however, illustrate a sort of \emph{reverse bayesianism} \citep{karni_reverse_2013}, where the hypothesis space itself expands over time.
	
	Instead of overcoming unawareness of states and actions, we could just represent unawareness as an infinite number of hidden states by modelling the problem as an \emph{infinite partially observable markov chain decision process} (i\pomdp) \citep{doshi-velez_infinite_2009}. This approach has several drawbacks. First, i\pomdp{}s don't currently address what to do when an unforeseen \emph{action} is discovered. More importantly, since the hidden variables are not tied to grounded concepts with explicit meaning, it is difficult for an agent to justify its decisions to a user, or to articulate queries about its current understanding of the world so as to solicit help from an expert.
	
	We instead propose a system where an agent makes explicit attempts to overcome its unawareness while constructing an interpretable model of its environment. This paper makes three contributions: First, an algorithm which incrementally learns \emph{all} components of an \fmdp{}. This includes the transition, reward, and value functions, but also the \emph{set of actions and belief variables themselves} (Section \ref{sec:unawarness}). Second, an expert-agent communication protocol (Section \ref{sec:expert-guidance})  which interleaves contextual advice with learning, and guarantees our agent converges to near-optimal behaviour, despite beginning unaware of factors critical to success. Third, experiments on small and large sequential decision problems showing our agent successfully learns optimal behaviour in practice (Section \ref{sec:experiments}).
	
	\section{The Learning Task}
	\label{sec:learning-task}
	
	We focus on learning \emph{episodic, finite state} \fmdp{}s with discrete states and actions. We begin with the formalisms for learning optimal behaviour in \fmdp{}s where the agent is fully aware of all possible states and actions. We then extend the task to one where the agent starts unaware of relevant variables and actions, and show how the agent overcomes this unawareness with expert aid.
	
	\subsection{Episodic Markov Decision Processes}
	\label{sec:emdps}
	An \mdp{} is a tuple $\langle \mathcal{S}, \mathcal{S}_s, \mathcal{S}_e, A, \mathcal{T}, \mathcal{R} \rangle$, where $\mathcal{S}$ and $A$ are the set of states and actions; $\mathcal{S}_{s}, \mathcal{S}_{e} \subseteq \mathcal{S}$ are possible start and end (terminal) states of an episode;  $\mathcal{T} : S \times A \times S \rightarrow [0,1]$ is the \emph{markovian transition function} $P(s' | s, a)$, and $\mathcal{R} : S \rightarrow \mathds{R}$ is the immediate reward function.\footnote{In this paper, we assume the agent's preferences depend only on the current state, and are both \emph{deterministic} and \emph{stationary}. Other works allow $\mathcal{R}$ to depend on the action and/or resulting state (i.e., $\mathcal{R} : S \times A \times S \rightarrow \mathds{R}$).} A policy $\pi : \mathcal{S} \times A \rightarrow [0,1]$ gives the probability $\pi(s,a)$ that an agent will take action $a$ in state $s$. When referring to the local time $m$ in episode $n$, we denote the current state and reward by $s_{m,n}$ and $r_{m,n} = \mathcal{R}(s_{m,n})$. When referring to the \emph{global time} $t$ across episodes, we denote them by $s_t$ and $r_t$. 
	
	The \emph{discounted return} for episode $n$ is: $G^n = \sum_{i=0}^{T} \gamma^i * r_{i,n}$, where $0 \leq \gamma \leq 1$ is the \emph{discount factor} governing how strongly the agent prefers immediate rewards. The agent's goal is to learn the \emph{optimal policy} $\pi_+$, which maximizes the expected discounted return in all states. The \emph{value function} $V_{\pi}(s)$ defines the expected return when following a given policy $\pi$, while the related action-value function $Q_{\pi}(s,a)$ gives the expected return of taking action $a$ in state $s$, and thereafter following $\pi$.
	
	\begin{gather}
	\label{eqn:value-function}
	V_\pi(s) = \mathcal{R}(s) + \gamma \sum_{s' \in S} P(s' | s, \pi(s)) V_{\pi}(s') \\
	\label{eqn:q-func}
	Q_\pi(s,a) = \mathcal{R}(s) + \gamma  \sum_{s' \in S} P(s' | s, a) V_\pi(s')
	\end{gather}
	
	If $\mathcal{T}$ and $\mathcal{R}$ are known, we can compute $\pi_+$ via \emph{value iteration} \citep{sutton_reinforcement_1998}. Further, we can measure the expected loss in discounted return of following policy $\pi$ versus $\pi_+$ using (\ref{eqn:policy-error}), which we refer to as the \emph{policy error}. If the agent's policy is unknown, we can approximate the policy error using (\ref{eqn:policy-error-approx}):

	\begin{gather}
	\label{eqn:policy-error}
	\mathit{Err}(\pi) = \sum_{s_0 \in \mathcal{S}_s} P(s_0) (V_{\pi_+}(s_0) - V_{\pi}(s_0)) \\
	\label{eqn:policy-error-approx}
	\mathit{Err}(t, t+k) =
	(\sum_{s_0 \in \mathcal{S}_s} P(s_0) V_{\pi_+}(s))
	- \frac
	{\sum_{i=t}^{t+k} G^i}
	{k}
	\end{gather}
	
	If all episodes eventually terminate, then (\ref{eqn:policy-error-approx}) will converge to (\ref{eqn:policy-error}). If our agent is \emph{$\epsilon$-greedy} (that is, in all states, has probability $\epsilon > 0$ of executing any action from $A$ at random), then termination in most \mdp{}s is guaranteed:
	
	\begin{definition}[Proper Policy]
		A policy $\pi$ is proper if, from all states $s \in \mathcal{S}$, acting according to $\pi$ guarantees one eventually reaches some terminal state $s' \in \mathcal{S}_{e}$.
	\end{definition}
		
	\begin{lemma}
		\label{thm:greed-proper}
		If an \mdp{} has a proper policy $\pi$, then any policy which is $\epsilon$-greedy with respect to $A$ is also proper.
	\end{lemma}
	
	\subsection{Learning FMDPs when Fully Aware}
	\label{sec:fmdps}
	
	If $\mathcal{T}$ or $\mathcal{R}$ are unknown, the agent must learn them using the data $D_{0:t} = [d_0, \dots, d_t]$ gathered from domain interactions. At time $t$, the \emph{sequential trial} $d_t = \langle s_t, a_t, s_{t+1}, r_{t+1} \rangle$ gives the current state $s_t$, action $a_t$, resulting state $s_{t+1}$ and the reward $r_{t+1}$ given on entering $s_{t+1}$. \fmdp{}s allow one to learn $\mathcal{T}$ for large \mdp{}s by representing states as a \emph{joint assignment} to a set of variables $\mathcal{X} = \{ X_1, \dots , X_n \}$ (Written as $\mathcal{S} = v(\mathcal{X})$). Similarly, the reward function is defined as a function $\mathcal{R} : v(scope(\mathcal{R})) \rightarrow \mathds{R}$, where $scope(\mathcal{R}) \subseteq \mathcal{X}$ are variables which determine the reward received in each state. To exploit conditional independence, $\mathcal{T}$ is then represented by a \emph{Dynamic Bayesian Network} (\dbn{}) \citep{dean_model_1989} for each action. That is, $\mathcal{T} = \{ dbn_{a_1}, \dots dbn_{a_n}\}$, where $dbn_{a} = \langle \Pa^a, \theta_a \rangle$. Here, $\Pa^a$ is a directed acyclic graph with nodes $\{ X_1, \dots, X_n,X'_1, \dots X'_n \}$ where, as is standard, node $X_i$ denotes the value of variable $X_i \in \mathcal{X}$ at the current time, while $X'_i$ denotes the same variable in the next time step. For each $X'_i$, $\Pa^a_{X'_i}$ defines the parents of $X'_i$. These are the only variables on which the value of $X'_i$ depends. We also make the common assumption that our {\sc dbn}s contain no \emph{synchronic arcs} \citep{degris_factored_2010}, meaning $\forall X'_i\ \forall a, \Pa^a_{X'_i} \subseteq \{ X_1, \dots, X_n \}$.
	
	This structure, along with the associated parameters $\theta_a$, allow us to write transition probabilities as a product of independent factors: $P(s' | s, a) = \prod_{X \in \mathcal{X}} \theta^a_{s'[X] , s[\Pa^a_{X'}]}$.
	Here, $s[\vec{Y}]$ is the projection of $s$ onto the variables in $\vec{Y}$, and $\theta^a_{X'=i,\Pa^a_{X'} = j}$ denotes the probability of variable $X$ taking on value $i$ given that the agent performs action $a$ when the variables $\Pa^a_{X'}$ have assignment $j$ in the current time step.\footnote{If the context is clear, we condense this notation to $\theta^a_{i,j}$.}
	
	Exploiting independence among belief variables doesn't guarantee a compact representation of $V_\pi$. We must also exploit the \emph{context-specific} independencies between assignments by representing $\mathcal{T}$ and $\mathcal{R}$ as \emph{decision trees}, rather than tables of values.
	
\begin{figure}
		\begin{subfigure}[b]{0.49\linewidth}
			\centering
			\includegraphics[scale=0.5]{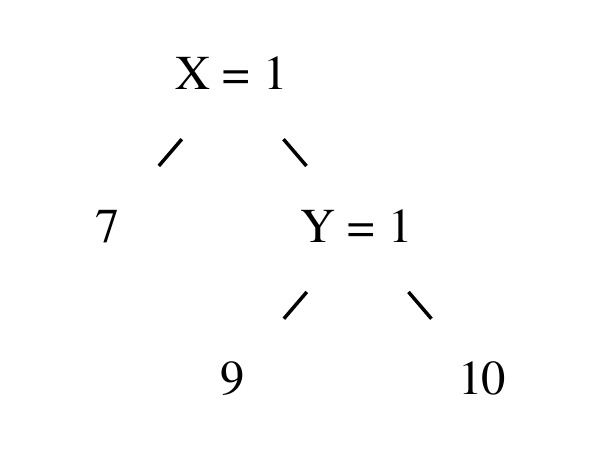}
			\caption{Reward}
			\label{fig:dt:reward}
		\end{subfigure}
		\begin{subfigure}[b]{0.49\linewidth}
			\centering
			\includegraphics[width=\textwidth]{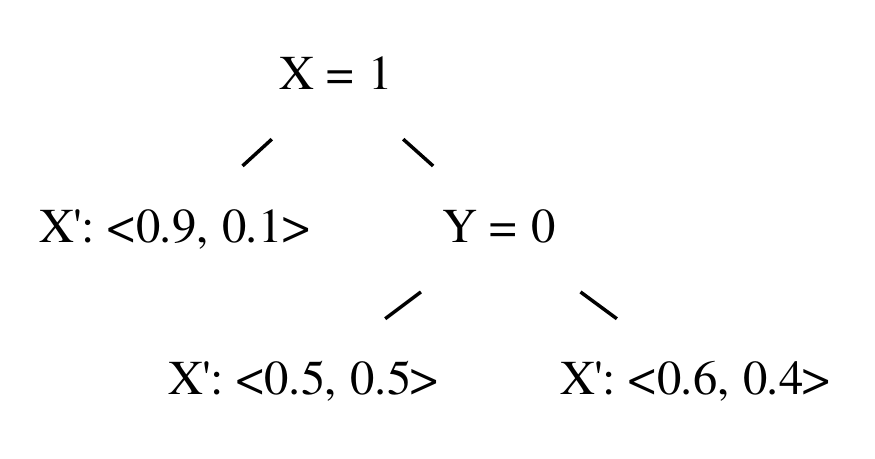}
			\caption{Conditional Probability}
			\label{fig:dt:cpt}
		\end{subfigure}
		
		\caption{Example decision trees}
		\label{fig:dt}
	\end{figure}

	Figure \ref{fig:dt} shows an example decision tree for $\mathcal{R}$ and $P(X' | X, Y)$. The \emph{leaves} are either rewards, or a distribution over the values of $X'$. The non-leaves are \emph{test nodes}, which perform a binary test of the form $(X = i?)$ to check whether variable $X \in \mathcal{X}$ takes on the value $i$ in the current state. Notice that when $X = 1$ is true, the distribution over $X'$ is conditionally independent of $Y$.
	
	Given trials $D_{0:t}$, we can estimate the most likely \dbn{} structure, tree structure and parameters, then subsequently estimate $V_{\pi_+}$ via a series of steps. Equation (\ref{eqn:bde-score}) is the \emph{Bayesian Dirichlet-Equivalent Score} \citep{heckerman_learning_1995}, which estimates the posterior probability $P(\Pa^a_{X'} | D_{0:t})$ that the true parents of $X'$ are $\Pa^a_{X'}$ by first integrating out all the possible parameters of the local probability distribution.
	
	\begin{equation}
	\label{eqn:bde-score}
	\begin{aligned}
	&P(\Pa^a_{X'} | D_{0:t}) \propto P(\Pa^a_{X'})\int_{\theta} P(D_{0:t} |\theta) P(\theta | \Pa^a_{X'}) \\
	&= P(\Pa^a_{X'})
	\prod_{\mathclap{j \in v(\Pa^a_{X'})}}
	\frac
		{\beta(N^a_{1,j} + \alpha^a_{1,j}, \dots, N^a_{m,j} + \alpha^a_{m,j})}
		{\beta(\alpha^a_{1,j}, \dots, \alpha^a_{m,j})}
	\end{aligned}
	\end{equation}
	
	Here, $N^a_{X=i,\Pa^a_{X'} = j}$ denotes the number of trials in $D_{0:t}$ in which action $a$ was taken in a state where the joint assignment to $\Pa^a_{X'}$ was $j$, resulting in a state where $X$ has the value $i$. The $\alpha$ terms are the hyper-parameters from the prior dirichlet distribution over parameters, and act as ``pseudo-counts'' when data is sparse. The prior $P(\Pa^a_{X'})$ is typically chosen to favour simple structures:
		
	\begin{equation}
	\label{eqn:struct-prior}
	P(\Pa^a_{X'}) = \rho^{|\Pa^a_{X'}|} (1 - \rho)^{|\mathcal{X}| - |\Pa^a_{X'}|}
	\end{equation}
	
	Equation (\ref{eqn:struct-prior}) penalizes \dbn{}s with many dependencies by attaching a cost $\rho < 0.5$ for each parent in $\Pa^a_{X'}$. Note, if the space of possible \dbn{}s is too large, we can restrict the parent sets considered reasonable by using common pruning heuristics or, for example, restricting the maximum in-degree.
	
	Given $\Pa^a_{X'}$, we can then compute each variable's most likely conditional probability tree structure $\textsc{dt}^a_{X}$, restricting node tests to members of $\Pa^a_{X'}$:
	
	\begin{gather}
	\label{eqn:dt-posterior}
	P(\textsc{dt}^a_{X} | \Pa^a_{X'}, D_{0:t}) \propto
	P(\textsc{dt}^a_{X}) P(D_{0:t} | \textsc{dt}^a_{X}) \\
	\label{eqn:dt-likelihood}
	P(D_{0:t} | \textsc{dt}^a_{X}) =
	\prod_{\mathclap{\ell \in Leaves(\textsc{dt}^a_{X})}}
	\frac
	{\beta(N^a_{1 | \ell} + \alpha^a_{1 | \ell} ,\ \dots\ ,N^a_{m | \ell} + \alpha^a_{m | \ell})}
	{\beta(\alpha^a_{1 | \ell},\ \dots\ ,\alpha^a_{m | \ell})}
	\end{gather}
	
	Rather than evaluating the probabilities of all possible $\textsc{dt}$ structures each step, we can incrementally update the most likely $\textsc{dt}$ as new trials arrive using \emph{incremental tree induction} (\iti{}), as described in \cite{utgoff_decision_1997}. While we lack the space to describe \iti{} in detail here, the algorithm broadly works by maintaining a single most-likely tree structure, with counts for all potential test assignments cached at intermediate nodes. As new trials arrive, the counts at relevant nodes become ``stale'', as there might now exist an alternative test which could replace the current one, resulting in a higher value for equation (\ref{eqn:dt-posterior}). If such a superior test exists, the test at this node is replaced, and the tree structure is transposed to reflect this change. We can also use \iti{} to learn a tree structure for $\mathcal{R}$ based on the trials seen so far. The only difference is that we use an information-gain metric to decide on the best test nodes, rather than (\ref{eqn:dt-posterior}).
	
	Finally, given $\textsc{dt}^a_X$, we compute the most likely parameters at each leaf via (\ref{eqn:expected-params}), where $N^a_{.,j} = \sum_{i \in v(X)} N^a_{i,j}$:
	
	\begin{equation}
	\label{eqn:expected-params}
	\begin{aligned}
	\mathbb{E}(\theta^a_{X=i,\Pa^a_{X'} = j} | D_{0:t}, \textsc{dt}^a_{X}) 
	&= \frac{N^a_{i,j} + \alpha^a_{i,j}}{N^a_{. , j} + \alpha^a_{. , j}}
	\end{aligned}
	\end{equation}
	
	Once our agent has a transition and reward tree, we can then use \emph{structured value iteration} (\svi{}) \citep{boutilier_stochastic_2000}---a variant of value iteration which works with decision trees instead of tables---to compute a compact representation of $V_{\pi_+}$. Algorithm \ref{alg:iSVI} shows an outline of an incremental version of \svi{} (i\svi{}) \citep{degris_learning_2006}, which allows the agent to gradually update its beliefs about the optimal value function in response to incoming trials. The algorithm takes the current estimate of the reward and transition functions ($\mathcal{R}_t$ and $\mathcal{T}_t$), along with the previous estimate of the optimal value function ($V_{t-1}$), and combines them to produce a new estimate for each state-action function ($Q^a_{t}$), and value function ($V_t$). For further details about the merge and reduce functions used in \svi{}, consult \cite{boutilier_stochastic_2000}.
	
	\begin{algorithm}
		\caption{Incremental \svi{} \citep{degris_learning_2006}}
		\label{alg:iSVI}
		\begin{algorithmic}[1]
			\Function{IncSVI}{$\mathcal{R}_t$, $\mathcal{T}_t$, $V_{t-1}$}
			\State $\forall a \in A: Q^a_t \gets \Call{Regress}{V_{t-1}, dbn_a, \mathcal{R}_t}$  
			\State $V_t \gets \Call{Merge}{\{ Q^a_t : \forall a \in A \}}$ (using maximization as the combination function)
			\State \Return $\{ V_t, \{ \forall a \in A: Q^a_t \} \}$
			\EndFunction
		\end{algorithmic}
	\end{algorithm}
	
	This section took an \emph{encapsulated} approach to learning $\mathcal{T}$ (In contrast to a \emph{unified} one in e.g., \cite{degris_learning_2006}). This means we separate the task of finding an optimal \dbn{} structure from the task of learning each local $\textsc{dt}$ structure. Such an approach significantly reduces the space of \textsc{dt}s that must be considered, but more importantly, provides us with posterior \emph{distributions} $P(\Pa^a_{X'} | D_{0:t})$ over parent structures. We will use these posterior distributions in section \ref{sec:adapt-transition} to conserve information when adapting to new discoveries.
	
	\section{Overcoming Unawareness}
	\label{sec:unawarness}
	
	So far, we've assumed our agent was aware of all relevant belief variables in $\mathcal{X}$, all actions $A$, and all members of $scope(\mathcal{R})$. We now drop this assumption. From here onward we denote the true set of belief variables, actions, and reward scope as $\mathcal{X}^+$, $A^+$ and $scope_+(\mathcal{R})$, and the learner's awareness of them at $t$ as  $\mathcal{X}^t$, $A^t$, and $scope_t(\mathcal{R})$
	
	Suppose $X^+ = \{ X_1, X_2, X_3 \}$, $X^0 = \{ X_1 \}$, $A^+ = \{ a, a' \}$, $A^t = \{ a \}$. We assume the agent can't observe the value of variables it is unaware of. In the medical example from before, if $X_3$ corresponds to a particular gene, then we assume the agent cannot detect the presence or absence of that gene if it is unaware that it exists. Similarly, we assume the agent cannot perform an action it is unaware of.\footnote{This assumption, while reasonable, may always not hold (E.g., an agent may lean on a button while unaware it is part of the task).} As a consequence, at time $t=0$, the agent does not directly observe the true trial $d_0$, but rather $d_0[\mathcal{X}^0] = \langle s_0[\mathcal{X}^0], a_0, s_1[\mathcal{X}^0], r_0 \rangle$. The key point here is that awareness of those missing factors may be crucial to successfully learning an optimal policy. For example, the transition between observed states may not obey the \emph{markov property} unless $X_2$ is observed, the best action may depend upon whether $X_3$ is true, or the optimal policy may sometimes involve performing $a'$. The next sections aims to answer two main questions. First, by what mechanisms can an agent discover and overcome its own unawareness by asking for help? Second, when an agent discovers a new belief variable or action, how can they integrate it into their current model while conserving what they have learned from past experience?
	
	\subsection{Expert Guidance}
	\label{sec:expert-guidance}
	Our agent can expand its awareness via advice from an expert. Teacher-apprentice learning is common in the real world, as it allows learners to receive contextually relevant advice which may inform them of new concepts they would not otherwise encounter.
	
	This paper assumes the expert has full knowledge of the true \mdp{}, is cooperative, and infallible. Further, we abstract away the complexity of grounding natural language statements in a formal semantics and instead assume that the agent and expert communicate via a pre-specified formal language (though see e.g., \cite{zettlemoyer_online_2007} for work on this problem). We do not, however, assume the expert knows the agent's \emph{current beliefs} about the decision problem.
	
	As argued in the introduction, the goal is to provide a minimal set of communicative acts so that interaction between the agent and expert proceeds analogously to human teacher-apprentice interactions. Concretely, this means we want our system to have two properties. First, the expert should, for the most part, allow the agent the opportunity to learn by themselves, interjecting only when the agent is performing sufficiently poorly, or when the agent explicitly asks for advice. Secondly, following the gricean maxims of conversation \citep{grice_logic_1975}, the expert should provide \emph{non-exhaustive} answers to queries, giving just enough information to resolve the agent's current query. We want this because in real world tasks with human experts it may be impossible to explain all details of a problem due to the cognitive constraints of the expert or costs associated with communication.
	
	The next sections identify three types of advice whose combination guarantee the agent behaves optimally in the long run, regardless of initial awareness.
	
	\subsubsection{Better Action Advice}
	\label{sec:better-action-advice}
	
	If the expert sees the agent perform a sub-optimal action, it can tell the agent a better action it could have taken instead. For example: ``When it is raining, take your umbrella instead of your sun hat''. Our goal is to avoid incessantly interrupting the agent each time it makes a mistake, so we specify the following conditions for when the agent is performing sufficiently poorly to warrant correction: Let $t$ be the current (global) time step corresponding to the $m$th step in the $nth$ episode. Similarly, let $t'$, $m'$, $n'$ be the time the expert last uttered advice. When (\ref{eqn:last-spoken-interval}-\ref{eqn:better-action-available}) hold, the expert utters advice of the form (\ref{eqn:better-action-utterance}):
	
	\begin{gather}
	\label{eqn:last-spoken-interval}
		t - t' > \mu  \\
	\label{eqn:suboptimal-returns}
		\mathit{Err(n', n)} > \beta \vee m > \kappa \\
	\label{eqn:better-action-available}
		\exists a' \in A^+, Q_{\pi_+}(a', s_{m, n}) > Q_{\pi_+}(a_{m, n} , s_{m, n}) \\
	\label{eqn:better-action-utterance}
		Q_{\pi_+}(w^s_{m,n}, a') > Q_{\pi_+}(w^s_{m,n},a_{m,n})
	\end{gather}
	
	Equation (\ref{eqn:last-spoken-interval}) ensures some minimum time $\mu$ has passed since the expert last gave advice. Equation (\ref{eqn:suboptimal-returns}) ensures the expert won't interrupt unless its estimate of the agent's policy error is above some threshold $\beta$, or if the agent is unable to reach a terminal state after some reasonable bound $\kappa$ (which is required because the agent's unawareness of $A^+$ may mean its current $\epsilon$-greedy policy is not proper). If episode $n$ is unfinished, the expert estimates the expected return via the heuristic $G^n \approx \sum_{i = 0}^{m - 1} \gamma^i r_{i, n} + \gamma^m V_{\pi_+}(s_{m,n})$, i.e., we optimistically assume the agent will follow $\pi_+$ from now on. Taken together, $\mu, \kappa$ and $\beta$ describe the expert's \emph{tolerance} towards the agent's mistakes. Finally, (\ref{eqn:better-action-available}) ensures a better action $a'$ actually exists at this time step. 
	
	Equation (\ref{eqn:better-action-utterance}) is the expert's utterance, and the term $w^s_{m,n}$ in it requires explanation. On first thought, the expert should utter $Q_{\pi_+}(s_{m,n}, a') > Q_{\pi_+}(s_{m,n}, a_{m,n})$, explicitly stating the full description of $s_{m,n}$. However, remember that the agent's awareness, $\mathcal{X}^t$, may be a tiny subset of $\mathcal{X}^+$. Uttering such advice may involve enumerating a huge number of variables the agent is currently unaware of. This is exactly the type of exhaustive explanation we wish to avoid, since such an explanation may place a cognitive burden on the expert, or confuse a learner. Conversely, we could instead have our expert project its intended utterance onto only those variables $\mathcal{X}^e$ for which the expert has explicit evidence the agent is aware of them (i.e., utter: $Q_{\pi_+}(s_{m,n}[\mathcal{X}^e], a') > Q_{\pi_+}(s_{m,n}[\mathcal{X}^e], a_{m,n})$). This can be understood by the agent without being made aware of any new variables, but might violate our assumption that the expert is truthful. For example, if $\exists s', s'[\mathcal{X}^e] = s_{m,n}[\mathcal{X}^e]$, but $Q_{\pi_+}(s', a) > Q_{\pi_+}(s', a')$.
	
	The solution is to use a \emph{sense ambiguous term} $w^s$, whose \emph{intended} denotation is the true state $s$ (i.e $\llbracket w_s \rrbracket \in v(\mathcal{X}^+)$), but whose \emph{default} interpretation by the agent is $s[\mathcal{X}^t]$. In words, it is as if the expert says \emph{``In the last step, it would have been better to do $a'$ than $a_{m,n}$''}.
	
	Thus, by introducing ambiguity, the agent can interpret the advice in two ways. The first is as a \emph{partial description} of the true problem, which is \emph{monotonically} true regardless of what it learns in future. On hearing (\ref{eqn:better-action-utterance}), the agent adds (\ref{eqn:better-action-monotonic:exists-action}-\ref{eqn:better-action-monotoni:exists-better-Q}) to its knowledge:
	
	\begin{gather}
	\label{eqn:better-action-monotonic:exists-action}
	a' \in A^+ \\	
	\label{eqn:better-action-monotoni:exists-better-Q}
	\exists s, s[\mathcal{X}^t] = s_{m,n}[\mathcal{X}^t] \wedge Q_{\pi_+}(s, a') > Q_{\pi_+}(s, a_{m,n})
	\end{gather}
		
	Additionally however, the agent can choose to add its current \emph{default interpretation} of the advice to its accumulated knowledge:
	
	\begin{equation}
	\label{eqn:better-action-defeasible}
	Q_{\pi_+}(s[\mathcal{X}^t], a') > Q_{\pi_+}(s[\mathcal{X}^t], a)
	\end{equation}
	
	The agent can then act on the expert's advice directly by choosing $a'$ whenever $s[\mathcal{X}^t] = s_{m,n}[\mathcal{X}^t]$, regardless of what seems likely from $D_{0:t}$. We can see that even with a cooperative and infallible expert, even abstracting away issues of grounding natural language, misunderstandings can still happen due to differences in agent and expert awareness. As the next section shows, such misunderstandings can reveal gaps in the agent's awareness and help to articulate queries whose answers guarantee the agent expands its awareness.
	
	Lemma \ref{thm:action-awareness-proper-policy} guarantees the expert's advice strategy reveals unforeseen actions to the agent so long as its performance in trials exceeds the expert's tolerance.\footnote{Proofs of  lemmas/theorems are in the technical supplement}
	
	\begin{lemma}
		\label{thm:action-awareness-proper-policy}
		Consider an \fmdp{} where $\pi_+$ is proper, an agent with awareness $\mathcal{X}^t \subseteq \mathcal{X}^+, A^t \subset A^+$, and expert acting with respect to (\ref{eqn:last-spoken-interval}-\ref{eqn:better-action-utterance}). If $\exists a \in image(\pi_+), a \notin A^t$ then as $k \rightarrow \infty$, either $Err(t, t + k) \rightarrow c$ with $c \leq \beta$ or the expert utters (\ref{eqn:better-action-available}) such that $a' \notin A^t$.
	\end{lemma}
	
	\subsubsection{Resolving Misunderstandings}
	\label{sec:resolving-misunderstandings}
	
	We noted before that the agent's defeasible interpretation of expert advice could result in misunderstandings. To illustrate, suppose the agent receives advice (\ref{eqn:advice-conflict-1}) and (\ref{eqn:advice-conflict-2}) at times $t -k$ and $t$:
	
	\begin{gather}
	\label{eqn:advice-conflict-1}
	Q_{\pi_+}(w^s_{t-k}, a) > Q_{\pi_+}(w^s_{t-k}, a') \\
	\label{eqn:advice-conflict-2}
	Q_{\pi_+}(w^s_t, a) < Q_{\pi_+}(w^s_t, a')
	\end{gather}
	
	While the intended meaning of each statement is true, the agent's default interpretations of $w^s_{t-k}$ and $w^s_{t}$ may be identical. That is, $s_{t-k}[\mathcal{X}^t] = s_{t}[\mathcal{X}^t]$. From the agent's perspective, (\ref{eqn:advice-conflict-1}) and (\ref{eqn:advice-conflict-2}) conflict, and thus give the agent a clue that its current awareness of $\mathcal{X}^+$ is deficient. To resolve this conflict, the agent asks (\ref{eqn:misunderstanding-q}) (in words, \emph{``which $X$ has distinct values in $s_{t-k}$ and $s_t$?''}) and receives an answer of the form (\ref{eqn:misunderstanding-a}):
		
	\begin{gather}
	\label{eqn:misunderstanding-q}
		?\lambda X(X\in \mathcal{X}^+ \wedge s_{t - k}[X] \neq s_{t}[X]) \\
	\label{eqn:misunderstanding-a}
	X \in \mathcal{X}^+
	\end{gather}
	
	Notice there may be multiple variables in $\mathcal{X}^+ \setminus \mathcal{X}^t$ whose assignments differ in $s_{t-k}$ and $s_t$. Thus, the expert's answer can be \emph{non-exhaustive}, providing the minimum amount of information to resolve the agent's conflict without necessarily explaining all components of the task. This means the agent must abandon its previous defeasible interpretation of (\ref{eqn:better-action-defeasible}), but can keep (\ref{eqn:better-action-monotonic:exists-action}-\ref{eqn:better-action-monotoni:exists-better-Q}), as these are true regardless of known variables. Lemma \ref{theorem:state-space-awareness} guarantees the expert will reveal new belief variables, provided such misunderstandings can still arise.
	
	\begin{lemma}
		\label{theorem:state-space-awareness}
		Consider an \fmdp{} where $\pi_+$ is proper and an agent with awareness $\mathcal{X}^t \subset \mathcal{X}^+, image(\pi_+) \subseteq A^t \subseteq A^+$. If $\exists s \exists s' \neq s, s[\mathcal{X}^t] = s'[\mathcal{X}^t]$, and $\pi_+(s) \neq  \pi_+(s')$, then as $k \rightarrow \infty$, either $Err(t, t + k) \rightarrow c$ $(c \leq \beta)$, or the expert utters (\ref{eqn:misunderstanding-a}) such that $X \notin \mathcal{X}^t$
	\end{lemma}
		
	\subsubsection{Unexpected Rewards}
	\label{sec:unexpected-rewards}
	
	In typical \fmdp{}s (where the agent is assumed fully aware of $\mathcal{X}^+, A^+$, and $scope_+(\mathcal{R})$), we tend only to think of the trials as providing \emph{counts}, but for an unaware agent, a trial $d_t = \langle s_t, a_t, s_{t+1}, r_{t+1} \rangle$ also encodes \emph{monotonic information}:
	
	\begin{equation}
	\label{eqn:trial-monotonic}
	\exists s, s[\mathcal{X}^t] = s_{t+1} \wedge \mathcal{R}_+(s) = r_{t+1}
	\end{equation}
	
	This constrains the form of $\mathcal{R}$ the agent must learn. Recall that $scope_t(\mathcal{R})$, may be only a subset $scope_+(\mathcal{R})$, so it might be impossible to construct an $\mathcal{R} : v(scope_t(\mathcal{R})) \rightarrow \mathds{R}$ satisfying all descriptions (\ref{eqn:trial-monotonic}) gathered so far. Further, those extra variables in $scope_+(\mathcal{R}) \setminus scope_t(\mathcal{R})$ may not be in $\mathcal{X}^t$. To resolve this, if the agent fails to construct a valid reward function, it asks (\ref{eqn:unforeseen-reward-q}) (in words, \emph{``which variable X (that I don't already know) is in $scope(\mathcal{R})$?''}), receiving an answer (\ref{eqn:unforeseen-reward-monotonic}):
	
	\begin{gather}
	\label{eqn:unforeseen-reward-q}
	?\lambda X(X\in \mathit{scope}_+(\mathcal{R}) \displaystyle\bigwedge_{X'\in scope_t(\mathcal{R})}
	X\neq X') \\
	\label{eqn:unforeseen-reward-monotonic}
		X \in scope_+(\mathcal{R}) \wedge X \in \mathcal{X}^+
	\end{gather}
		
	Again, the agent may be unaware of many variables in $scope_+(\mathcal{R})$, so (\ref{eqn:unforeseen-reward-monotonic}) may be \emph{non exhaustive}. Even so, we can guarantee that the agent's learned reward function eventually equals $\mathcal{R}_+$:
	
	\begin{lemma}
		\label{thm:reward-scope-awareness}
		Consider an \fmdp{} where $\pi_+$ is proper and an agent with awareness $A^t \subseteq A^+$, $\mathcal{X}^t \subseteq \mathcal{X}^+$, $scope_t(\mathcal{R}) \subseteq scope_+(\mathcal{R})$. As $k \rightarrow \infty$, there exists a $K$ such that for all $k \geq K$, $\mathcal{R}_{t+k}(s) = \mathcal{R}_{+}(s)$ for all states $s$ reachable using $A^t$.
	\end{lemma}
	
	\subsection{Adapting the Transition Function}
	\label{sec:adapt-transition}
	
	Section \ref{sec:expert-guidance} showed three ways the agent could expand its awareness of $\mathcal{X}$, $A$, and $scope(\mathcal{R})$. If we wish to improve on the naive approach of restarting learning when faced with such expansions, we must now specify how the agent adapts $\mathcal{T}$ and $\mathcal{R}$ to such discoveries.
	
	Adapting $\mathcal{T}$ upon discovering a new action $a'$  at time $t$ is simple: Since the agent hasn't performed $a'$ in any previous trial, it can just create a new \dbn{}, $dbn_{a'}$, using the priors outlined in section \ref{sec:fmdps}. Our new model at time $t$ then becomes $\mathcal{T} = \{ dbn^t_{a_1}, \dots dbn^t_{a_n} \} \cup \{dbn_{a'}\}$.
	
	The more difficult issue is adapting $\mathcal{T}$ upon discovering a new belief variable $Z$. The main problem is that the agent's current distributions over \dbn{}s no longer cover all possible parent sets for each variable, nor all \textsc{dt}s. For example, the current distribution over $\Pa^a_{X'}$ does not include the possibility that $Z$ is a parent of $X'$. Worse, since we assume in general that the agent \emph{cannot observe $Z$'s past values} in $D_{0:t}$, it cannot observe the true value of $N^a_{Z  = i| j}$, nor $N^{a}_{X = i | \Pa^a_{X'} = j}$ when $Z \in \Pa^a_{X'}$. The $\alpha$-parameters involving $Z$ are also undefined, yet we need them to calculate structure probabilities (\ref{eqn:bde-score}, \ref{eqn:dt-posterior}) and parameters via (\ref{eqn:expected-params}).
	
	The problem is that new variables make the size of each (observed) state \emph{dynamic}, in contrast to standard problems where they are \emph{static} (e.g., $\langle X_1 = 0, X_2 = 1 \rangle$ becomes $\langle X_1 = 0, X_2 = 1, Z = ?\rangle$ ) . We could phrase this as a missing data problem: $Z$ was hidden in the past but visible in future states, so treat the problem as a \pomdp{} and estimate missing values via e.g., \emph{expectation maximization} \citep{friedman_bayesian_1998}. However, such methods commit us to costly passes over the full state-action history, and make it hard to learn $\textsc{dt}$ structures with enough sparseness to ensure a compact value function. Alternatively, we could ignore states with missing information when counts involving $Z$ are required. For example, we could use $P(\Pa^a_{X'} | D_{t:n})$ to score $\Pa^a_{X'}$ when $Z \in \Pa^a_{X'}$ but use $P(\Pa^a_{X'} | D_{0:n})$ when $Z \notin \Pa^a_{X'}$. However, as \cite{friedman_sequential_1997} points out, most structure scores, including (\ref{eqn:bde-score}), assume we evaluate models with respect to the \emph{same data}. If two models are compared using different data sets (even if they come from the same underlying distribution), the learner tends to favour the model evaluated with the smaller amount of data. Instead, our method discards the data gathered during the learner's previous deficient view of the hypothesis space, but \emph{conserves} the relative  \emph{posterior} probabilities learned from past data to construct new \emph{priors} for the $\Pa^a$, $\textsc{dt}^a$ and $\theta^a$ in the expanded belief space.
	
	\subsubsection{Parent Set Priors}
	
	On discovering $Z$, the agent must update $P(\Pa^a_{X'})$ for each $X \neq Z$ and $a \in A^t$ to include parent sets containing $Z$. In (\ref{eqn:old-parent-sets-new-vocab}) we construct a new prior $P'(\Pa^a_{X'})$ using the old posterior:
	
	\begin{equation}
	\label{eqn:old-parent-sets-new-vocab}
	\begin{aligned}
	&P'(\Pa^a_{X'}) =
	\begin{cases}
	(1 - \rho) P(\Pa^a_{X'} | D_{0:t}) & \text{if } Z \notin \Pa^a_{X'} \\
	\rho P((\Pa^a_{X'} \setminus Z) | D_{0:t}) & \text{otherwise}
	\end{cases}
	\end{aligned}
	\end{equation}
	
	This preserves the likelihoods among the parent sets that do not include $Z$. It also maintains our bias towards simpler structures by re-assigning only a portion $\rho$ of the probability mass to parent sets including $Z$. To define $P(\Pa^a_{Z'})$---the distribution over parent sets for the newly discovered variable $Z$---we default to (\ref{eqn:struct-prior}), since the agent has no evidence (yet) concerning $Z$'s parents.
	
	\subsubsection{Decision Tree and Parameter Priors}
	
	We must also update $P(\textsc{dt}^a_{X} | \Pa^a_{X'})$, and $P(\theta^a_{X, \Pa^a_{X'}} | \Pa^a_{X'})$ to accommodate $Z$. Here, we return to the issue of the counts $N^a_{i | j}$ and the associated $\alpha$-parameters. As mentioned earlier, we wish to avoid the complexity of estimating $Z$'s past values. Instead, we \emph{throw away} the past counts of $N^a_{i | j}$, but \emph{retain} the relative likelihoods they gave rise to by packing these into new $\alpha$-parameters, as shown in (\ref{eqn:alpha-update}-\ref{eqn:tree-new-prior}):
	
	\begin{gather}
	\label{eqn:alpha-update}
	\alpha^a_{X = i | Y = j} := 
	\begin{cases}
	\frac{K}{|v(Z \cup Y)|} & \text{ if } X = Z \\
	\frac{K}{|v(Y)|} P(i, j[Y \setminus Z] \rvert dbn^t_a)  & \text{else}
	\end{cases} \\
	\label{eqn:tree-new-prior}
	P'(\textsc{dt}^a_{X'} | \Pa^a_{X'}) \propto
	\prod_{\mathclap{\ell \in leaves(\textsc{dt}^a_{X'})}}
	\beta(\alpha^a_{X' = 1 | \ell}, \dots\ , \alpha^a_{X'=n | \ell})
	\end{gather}
	
	Equation (\ref{eqn:alpha-update}) summarizes $D_{0:t}$ via inferences on the old best \dbns, then encodes these inferences in the new $\alpha$-parameters. The revised $\alpha$-parameters ensure the new tree structure prior and expected parameters defined via (\ref{eqn:tree-new-prior}) and (\ref{eqn:expected-params}) bias towards models the agent previously thought were likely. Indeed, the larger the (user-specified) $K$ parameter is, the more the distributions learned \emph{before} discovering $Z$ influence the agent's reasoning \emph{after} discovering $Z$.	
	
	\subsection{Adapting Reward and Value Trees}
	\label{sec:adapting-reward-value}
	
	On becoming aware $Z$ is part of $scope_+(\mathcal{R})$, the agent may wish to restructure its reward tree. This is because awareness that $Z \in scope_+(\mathcal{R})$ means there are tests of the form $Z=i$ that the agent has not yet tried which may produce a more compact tree. In the language of \iti{}, the current test nodes are ``stale'', and must be re-checked to see if a replacement test would yield a tree with better information gain. If the agent was unaware of $Z$ (i.e, $Z \notin \mathcal{X}^t$), we can still test on assignments to $Z$ by following the \iti{} convention that any state where $Z$ is missing automatically fails any test on $Z$.
	
	Once we have updated $\mathcal{T}$ and $\mathcal{R}$, there is no need to make further changes to  $V_t$ in response to a new action $a'$ or variable $Z$. In effect, this encodes our conservative intuition that the true $V_{\pi_+}$ is more likely to be closer to the agent's current estimate $V_t$ than some arbitrary value function. The agent essentially assumes (in absence of further information) that the value of a state is indifferent to this newly discovered factor. In subsequent trials where the agent performs $a'$ or observes $Z$, algorithm \ref{alg:iSVI} ensures information about this new factor is incorporated into the agent's value function.

	\begin{algorithm}
		\caption{Learning \fmdp{}s with Unawareness}
		\label{alg:full-system}
		\begin{algorithmic}[1]
			\Function{LearnFMDPU}{$A^0$, $\mathcal{X}^0$, $\mathcal{T}_0$, $Q_0$, $V_0$, $s_0$}
			\For{$t=1 \dots \mathit{maxTrials}$}
			\State $\langle s_{t}, r_{t} \rangle \gets$ \Call{$\epsilon$-greedy}{$s_{t-1}$, $Q_{t-1}$, $adv_{0:t-1}$}
			\State $\langle \mathcal{T}_t, \mathcal{R}_t \rangle \gets $ Add $\langle s_t, r_t \rangle$ via (\ref{eqn:bde-score}-\ref{eqn:expected-params}) \& \textsc{iti}
			\If{Update to $\mathcal{R}_t$ fails}
				\State $Z \gets $ Ask expert (\ref{eqn:misunderstanding-q})
				\State $\langle scope_t(\mathcal{R}), \mathcal{X}^t \rangle \gets$ Append $Z$ to each
				\State $\mathcal{R}_t \gets $ Update via \iti{}
			\EndIf
			\If{ (\ref{eqn:last-spoken-interval}-\ref{eqn:better-action-available}) are true}
				\State $\mathit{adv}_t \gets$ Expert advice of form (\ref{eqn:better-action-utterance})
				\If{$adv_{t}$ mentions action $a' \notin A^{t-1}$}
					\State $\mathcal{A}^t \gets \mathcal{A}^{t-1} \cup \{a'\}$
					\State $\mathcal{T}_t \gets \mathcal{T}_{t-1} \cup \{dbn_{a'}\}$ made via (\ref{eqn:struct-prior})
				\EndIf
				\If{$adv_{0:t-1}$ conflicts with $adv_t$}
					\State $Z \gets$ Ask expert (\ref{eqn:misunderstanding-q})
					\State $\mathcal{X}^t \gets \mathcal{X}^{t-1} \cup \{ Z \}$
				\EndIf
			\EndIf	
			\If{$\mathcal{X}^{t-1} \neq \mathcal{X}^t$}
				\State $\mathcal{T}_t \gets$ Update via (\ref{eqn:alpha-update}, \ref{eqn:bde-score}, \ref{eqn:tree-new-prior}, \ref{eqn:dt-posterior}, \ref{eqn:expected-params})
			\EndIf
			\State $\langle V_t, Q_t \rangle \gets $ \Call{IncSVI}{$\mathcal{R}_t, \mathcal{T}_t, \mathcal{V}_{t-1}$}
			\EndFor
			\EndFunction
		\end{algorithmic}
	\end{algorithm}
	
	Algorithm \ref{alg:full-system} outlines how the agent updates $\mathcal{T}$, $\mathcal{R}$, and $V$ in response to new data and expert advice. Given algorithm \ref{alg:full-system}, theorem \ref{thm:full-unawareness-guarantee} guarantees our agent behaves indistinguishably from a near-optimal policy in the long run, regardless of initial awareness (provided all $X \in \mathcal{X}^+$ are relevant to expressing the optimal policy).
	
	\begin{theorem}
		\label{thm:full-unawareness-guarantee}
		Consider an \fmdp{} where $\pi_+$ is proper and an agent with initial awareness $\mathcal{X}^0 \subseteq \mathcal{X}^+, A^0 \subseteq A^+$, and $scope_0(\mathcal{R}) \subseteq scope_+(\mathcal{R})$ acts according to algorithm \ref{alg:full-system}. If for all $X \in \mathcal{X}^+$, there exists a pair of states $s, s'$ such that $s[\mathcal{X}^+ \setminus X] = s'[\mathcal{X}^+ \setminus X]$, $s[X] \neq s'[X]$, and $\pi_+(s) \neq \pi_+(s')$, then as $t \rightarrow \infty$, $Err(0, t) \rightarrow c$ such that $c \leq \beta$
	\end{theorem}
	\section{Experiments and Results}
	\label{sec:experiments}
	
	Our experiments show that agents following algorithm \ref{alg:full-system} converge to near-optimal behaviour in both theory and practice. Further, we show that conserving information on $\mathcal{T}$ and $V$ gathered before each new discovery allows our agent learn faster than one which abandons this information. We do not investigate assigning an explicit budget to agent-expert communication, leaving this to future work. However we do show how varying the expert's tolerance affects the agent's performance.
		
	We test agents on two well-known problems: \emph{Coffee-Robot} and \emph{Factory}.\footnote{Full specifications at \url{https://cs.uwaterloo.ca/~jhoey/research/spudd/index.php}} In each, our agent begins with only partial awareness of $\mathcal{X}^+$, $A^+$ and $scope_+(\mathcal{R})$. The agent takes actions for $T$ time steps, using an $\epsilon$-greedy policy ($\epsilon = 0.1$). When the agent enters a terminal state, we reset it to one of the initial states randomly. We use the \emph{cumulative reward} across all trials as our evaluation metric, which acts as a proxy for the quality of the agent's policy over time. To make the results more readable, we apply a discount of $0.99$ at each step, resulting in the metric $R^{disc}_t = r_t + 0.99 * R^{disc}_{t-1}$.
	
	We test several variants of our agent to show the effectiveness of our approach. The \textbf{default} agent follows algorithm \ref{alg:full-system} as is, with parameters $\rho = 0.1$, $K = 5.0$, $\mu=10$, $\beta=0.1$, $\kappa = 50$ in equations (\ref{eqn:struct-prior}), (\ref{eqn:old-parent-sets-new-vocab}), (\ref{eqn:alpha-update}), and (\ref{eqn:last-spoken-interval}-\ref{eqn:better-action-available}) respectively. The \textbf{nonConservative} agent does not conserve information about $V$, nor $\mathcal{T}$ via (\ref{eqn:old-parent-sets-new-vocab}-\ref{eqn:tree-new-prior}) when a new factor is discovered. Instead, it resets $V$ and $\mathcal{T}$ to their initial values. This agent is included to show the value of conserving past information as $\mathcal{X}$ and $A$ expand. The \textbf{truePolicy} and \textbf{random} agents start with full knowledge of the true \fmdp{}, and execute an $\epsilon$-greedy version of $\pi_+$, or a choose random action respectively. These agents provide an upper/lower bound on performance. The \textbf{lowTolerance} / \textbf{highTolerance} agents change the expert's tolerance to $\beta=0.01$ and $\beta=0.5$.
	
	\subsection{Coffee Robot}
	\label{sec:coffee-robot}
	
	\emph{Coffee-Robot} is a small sequential problem where a robot must purchase coffee from a cafe, then return it to their owner. Also, the robot gets wet if it has no umbrella when it rains. The problem has 6 boolean variables---$\textsc{huc}$ (user has coffee), $\textsc{hrc}$ (robot has coffee), $\textsc{r}$ (raining), $\textsc{w}$ (wet), $\textsc{l}$ (location), $\textsc{u}$ (umbrella)--- and 4 actions---\textsc{move, delc, buyc} and \textsc{getu}--- making 256 state/action pairs. The terminal states are those where $\textsc{huc}=1$; initial states are all non-terminal ones. Our agent has initial awareness $A^0 = \{\textsc{move} \}$, $\mathcal{X}^0= scope_0(\mathcal{R}) = \{ \textsc{huc} \}$ and discount factor $\gamma = 0.8$.\footnote{Original setting was $\gamma=0.9$. Changed to make $\pi_+$ proper.}
	
	\begin{figure}
		\begin{subfigure}{\linewidth}
			\includegraphics[width=0.49\textwidth]{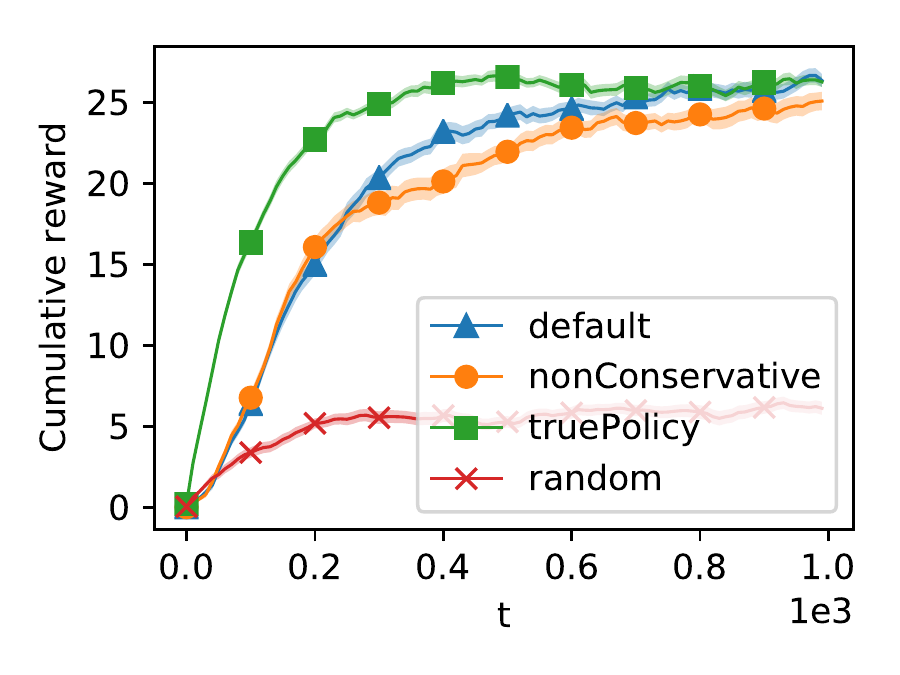}
			\includegraphics[width=0.49\textwidth]{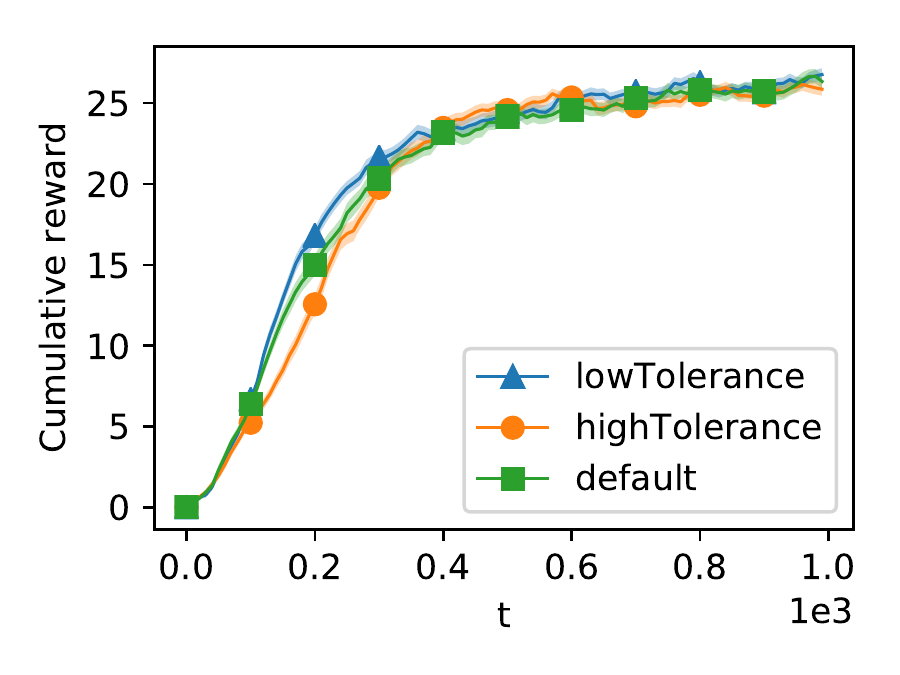}
			\caption{\emph{Coffee Robot} ($T=1000$. Average of 50 experiments)}
			\label{fig:coffee-rewards-agents}
		\end{subfigure}
		
		\begin{subfigure}{\linewidth}
			\includegraphics[width=0.49\textwidth]{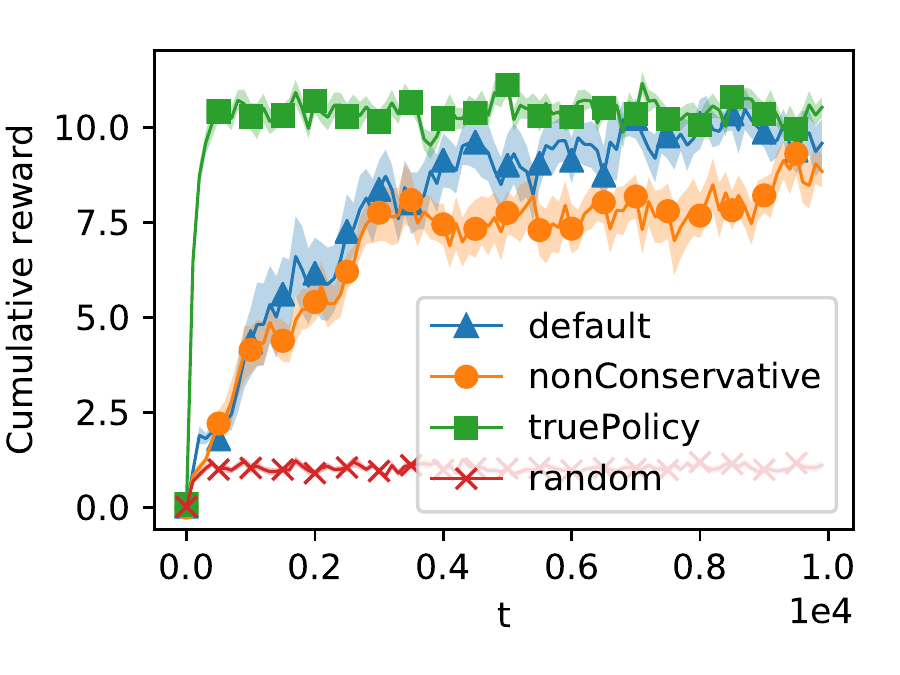}
			\includegraphics[width=0.49\textwidth]{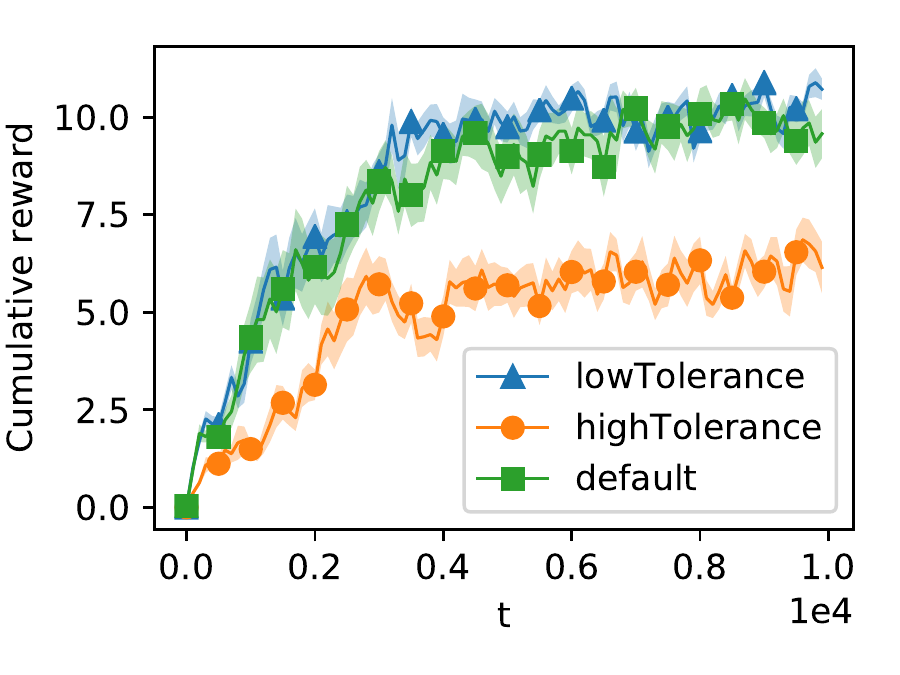}
			\caption{\emph{Factory} ($T=10000$. Average of 20 experiments)}
			\label{fig:factory-rewards}
		\end{subfigure}
		\caption{Cumulative Rewards. Shaded areas represent standard error from the mean.}
	\end{figure}
	
\begin{figure}
		\centering
		\begin{subfigure}[b]{0.59\linewidth}
			\centering
			\includegraphics[width=\textwidth]{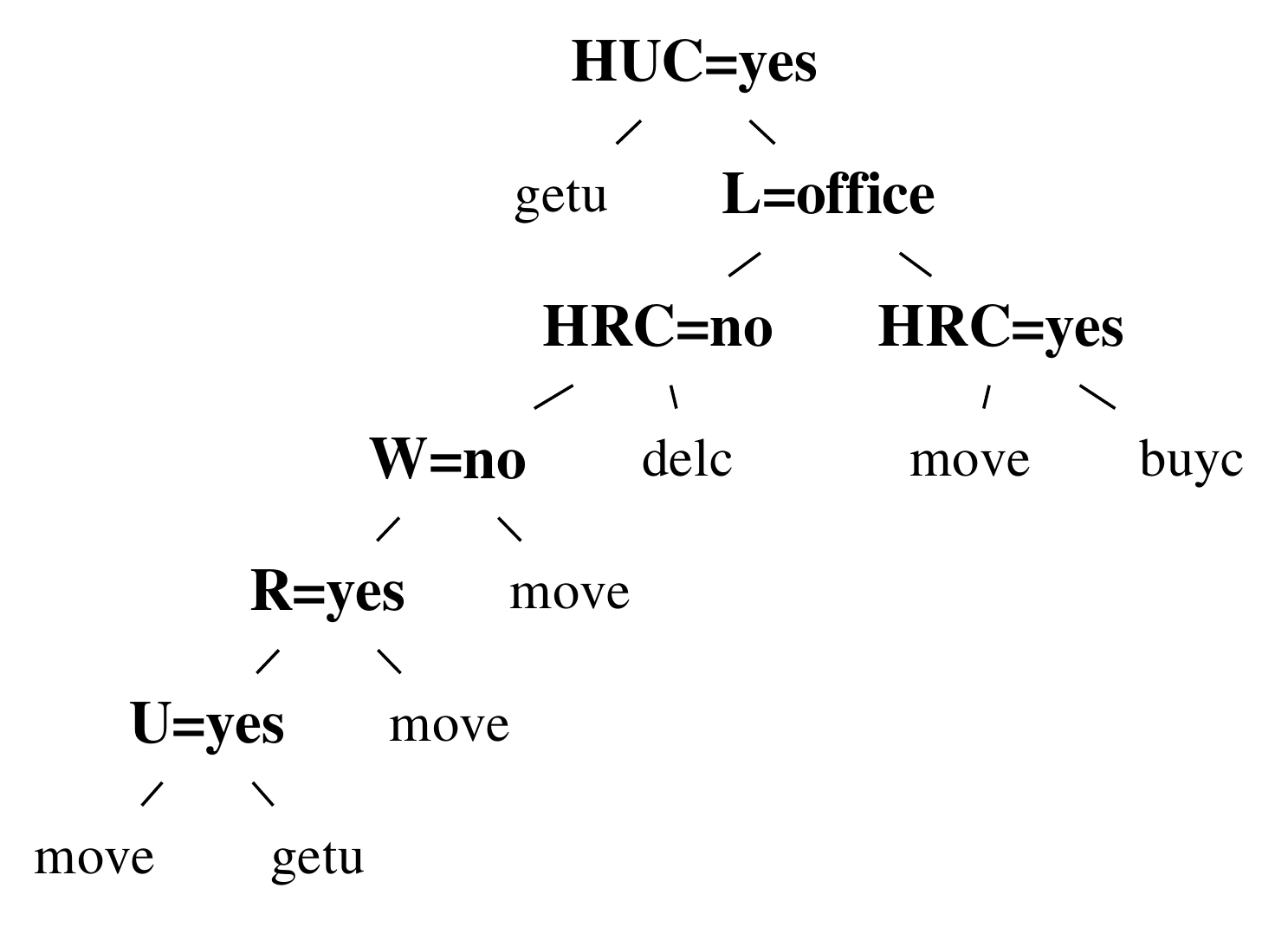}
			\caption{Default Tolerance}
		\end{subfigure}
		\begin{subfigure}[b]{0.39\linewidth}
			\centering
			\includegraphics[width=\textwidth]{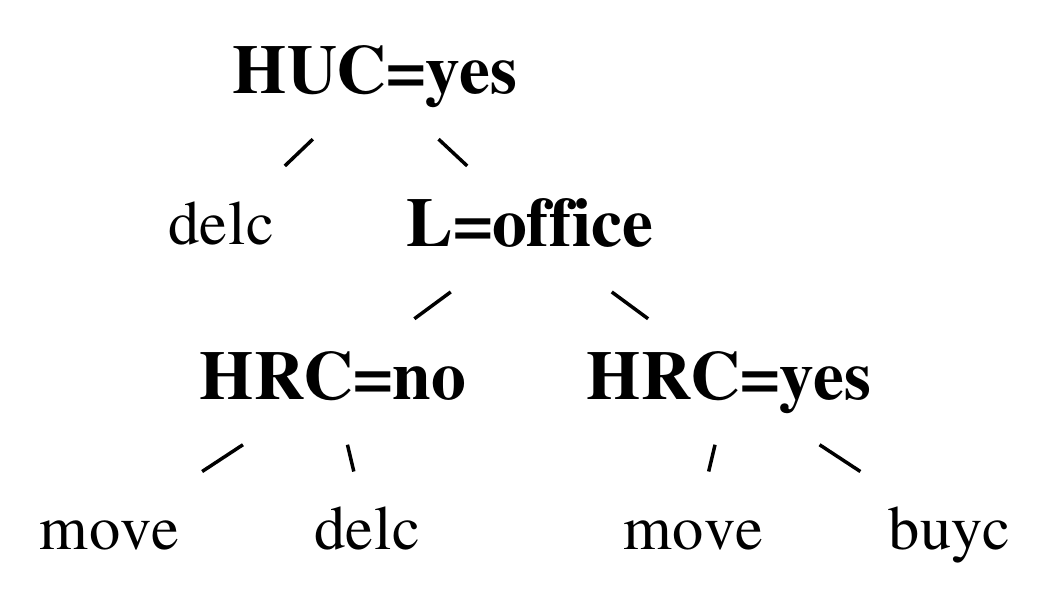}
			\caption{High Tolerance}
		\end{subfigure}
		\caption{Typical final policy depending on tolerance}
		\label{fig:coffee-final-policies}
	\end{figure}

\begin{figure}
		\centering
		\begin{subfigure}{0.49\linewidth}
			\includegraphics[width=\textwidth]{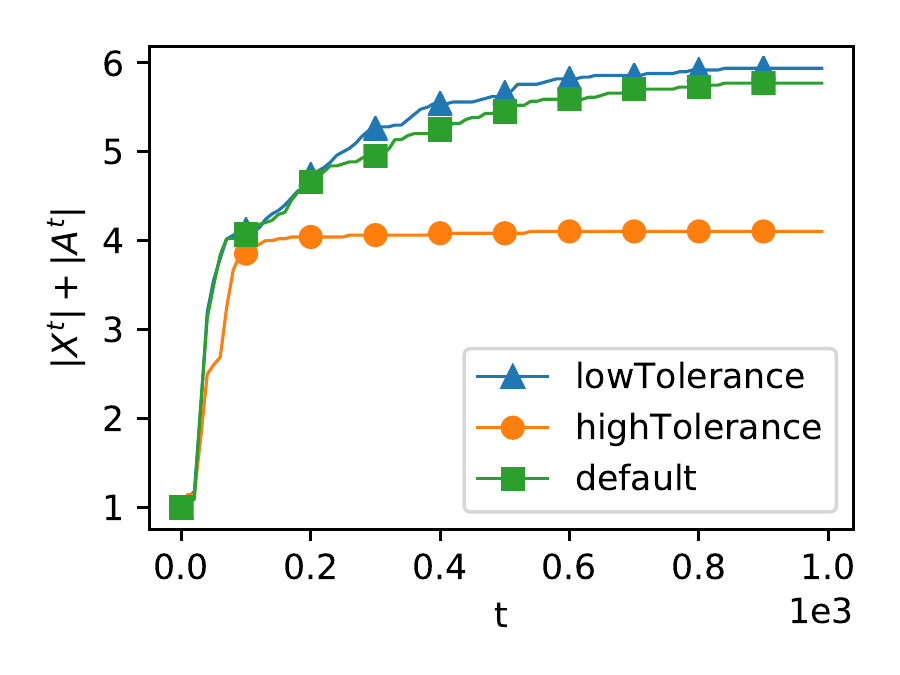}
			\caption{\emph{Coffee Robot} task}
		\end{subfigure}
		\begin{subfigure}{0.49\linewidth}
			\includegraphics[width=\textwidth]{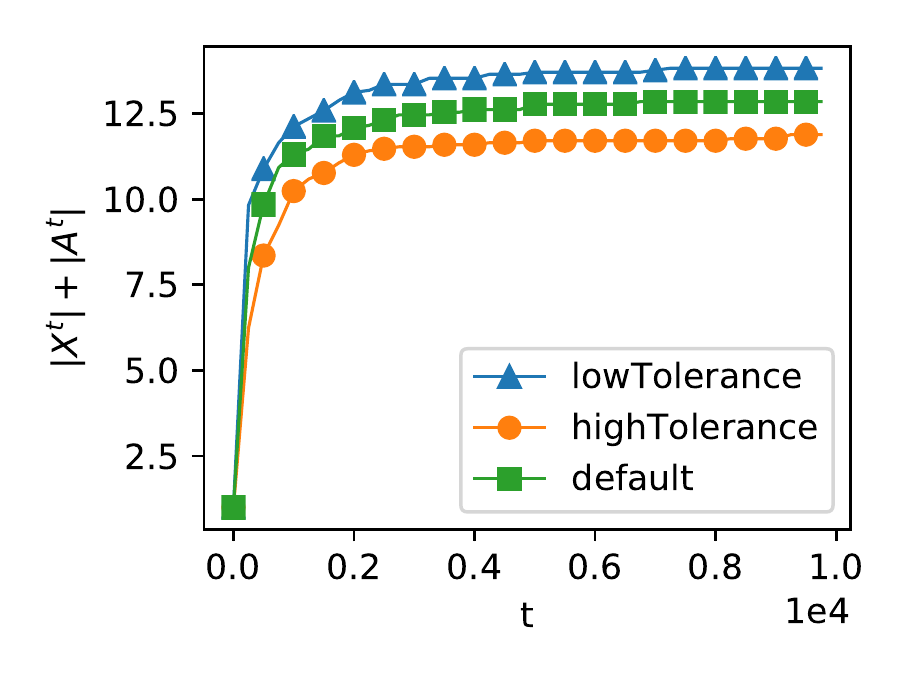}
			\caption{\emph{Factory} task}
		\end{subfigure}

		\caption{Awareness of $|\mathcal{X}^+|$ and $|A^+|$}
		\label{fig:coffee-vocab}
	\end{figure}

	Figure \ref{fig:coffee-rewards-agents} shows each agent's (discounted) cumulative reward. Despite starting unaware of factors critical to success, the default agent quickly discovers the relevant actions and beliefs with the expert's aid, and converges on the optimal policy. The non-conservative agent also learns the optimal policy, but takes longer. This shows the value of conserving $\mathcal{T}$ and $V$ on discovering new beliefs. We also see how expert tolerance affects performance. The agent paired with high tolerance expert learns a (marginally) worse final policy, but this makes little difference to cumulative reward. Figure \ref{fig:coffee-final-policies} shows why: The agent learned a ``good enough'' policy, so the expert doesn't reveal the ``get umbrella'' (\textsc{getu}) action, which yields only a minor increase in reward. Figure \ref{fig:coffee-vocab} supports this explanation, showing how more tolerant experts reveals less variables over time.
	
	\subsection{Factory}
	\label{sec:factory}
	
	\emph{Factory} is a larger problem ($|A^+| = 14$, $\mathcal{X}^+ = 14$, 774144 state/action pairs), which shows our method works on more realistically sized tasks. Here, an agent must shape, paint and connect two widgets to create products of varying quality. Some actions (like bolting) produce high quality products, whereas others (like gluing) produce low quality products. The agent receives a higher reward for producing goods which match the demanded quality.\footnote{Rewards were scaled to range 0.0-1.0 and, to make $\pi_+$ proper, terminal states which previously gave $0$ reward were given a small reward of $0.01$.}. The terminal states are those where $\textsc{connected}=1$; initial states are non-terminals where it is possible to connect two components. Our agent's initial awareness is $\mathcal{X}^0 = scope_0(\mathcal{R}) = \{ \textsc{connected} \}$, $A^0 = \{ \textsc{bolt, glue, drilla, drillb} \}$, with $\gamma=0.9$. This represents a simplified task where the agent thinks the only goal is connecting the widgets.
	
	Figure \ref{fig:factory-rewards} shows results similar to previous experiments. The default agent converges on optimal behaviour, and does so quicker than the non-conservative agent. Varying the expert's tolerance now has a larger effect on the rate at which factors are discovered and on convergence towards the optimal policy (presumably because there are many more unforeseen variables/actions the agent can discover in this larger problem).
	
	\section{Related Work}
	\label{sec:related-work}
	
	Models of unawareness exist in logic and game theory \citep{board_two_2011,heifetz_dynamic_2013,feinberg_games_2012}, but interpret (un)-awareness from an omniscient view. We instead model awareness from the agent's view and offer methods to \emph{overcome} one's own unawareness.
	
	\cite{rong_learning_2016} defines unawareness similarly to us, using \emph{markov decision processes with unawareness} ({\sc mdpu}s) to learn optimal behaviour when an agent starts unaware of some actions. They apply {\sc mdpu}s to a robotic-motion problem with around 1000 discretised atomic states. The agent uses an \emph{explore} move, which randomly reveals useful motions they were previously unaware of. Our work differs from theirs in several ways. First, we provide a concrete mechanism for discovering unforeseen factors via expert advice, rather than random discovery from the agent's own exploration. Second, we allow the agent to discover \emph{explicit belief variables} rather than atomic states, and focus more on exploiting the inherent structure in problems with a large number of features. This enables us to scale up to complex decision problems, where the agent converges on an optimal policies in a (true) state space around a million atomic states, as opposed to around 1000. \cite{mccallum_reinforcement_1996} also learn an increasingly complex representation of the state space by gradually distinguishing between states which yield different rewards. Rather than dealing with unawareness, their approach focusses on \emph{refining} an existing state space. In other words, they do not support introducing \emph{unforeseen} states or actions that the learner was unaware of before learning.
	
	Several works use expert interventions to improve performance via reward shaping and corrections \citep{stone_interactively_2009,torrey_teaching_2013}. Yet all such methods assume the expert's intended meaning can be understood without expanding the agent's current state and action space. Our work allows experts to utter advice where ambiguity arises from their greater awareness of the problem.
		
	\section{Conclusion}
	\label{sec:conclusion}
	
	We have presented an agent-expert framework for learning optimal behaviour in both small and large \fmdp{}s even when one starts unaware of factors critical to success. Further, we've shown that conserving one's beliefs helps improve the effectiveness of learning. In future work, we aim to lift some assumptions imposed on the expert, and expand the expressiveness of its advice. For instance, we could let the expert be fallible, or allow questions on the \emph{structure} of $\mathcal{T}$, as \cite{masegosa_interactive_2013} do for Bayesian Networks.
	
	\bibliography{thesis}
	
\end{document}